\documentclass[runningheads]{llncs}
\usepackage{caption}
\usepackage{makecell}
\usepackage{subcaption}
\usepackage[T1]{fontenc}
\usepackage{graphicx}
\usepackage{amsfonts}
\usepackage{amsmath}
\usepackage{booktabs}
\usepackage{multirow}
\begin{document}
%

\title{Diffusion Model based Semi-supervised Learning on Brain Hemorrhage Images for Efficient Midline Shift Quantification}
\titlerunning{Diffusion model based ICH midline shift quantification}
%
\author{Shizhan Gong\inst{1}, Cheng Chen\inst{1}, Yuqi Gong\inst{1}, Nga Yan Chan\inst{2}, Wenao Ma\inst{1}, Calvin Hoi-Kwan Mak\inst{3}, Jill Abrigo\inst{2}, Qi Dou\inst{1}}
\authorrunning{S.Z. Gong et al.}
%
\institute{$^1$Department of Computer Science and Engineering, \\The Chinese University of Hong Kong, Hong Kong, China\\
$^2$Department of Imaging and
Interventional Radiology, \\ The Chinese University of Hong Kong, Hong Kong, China\\
$^3$Queen Elizabeth Hospital, Hong Kong, China
}
\maketitle              
\begin{abstract}
Brain midline shift (MLS) is one of the most critical factors to be considered for clinical diagnosis and treatment decision-making for intracranial hemorrhage. 
Existing computational methods on MLS quantification not only require intensive labeling in millimeter-level measurement but also suffer from poor performance due to their dependence on specific landmarks or simplified anatomical assumptions. In this paper, we propose a novel semi-supervised framework to accurately measure the scale of MLS from head CT scans. We formulate the MLS measurement task as a deformation estimation problem and solve it using a few MLS slices with sparse labels. Meanwhile, with the help of diffusion models, we are able to use a great number of unlabeled MLS data and 2793 non-MLS cases for representation learning and regularization. The extracted representation reflects how the image is different from a non-MLS image and regularization serves an important role in the sparse-to-dense refinement of the deformation field. Our experiment on a real clinical brain hemorrhage dataset has achieved state-of-the-art performance and can generate interpretable deformation fields.

\end{abstract}
\section{Introduction}

\begin{figure}[t]
     \centering
     \begin{subfigure}[b]{0.24\textwidth}
         \centering
         \includegraphics[width=\textwidth, height=\textwidth]{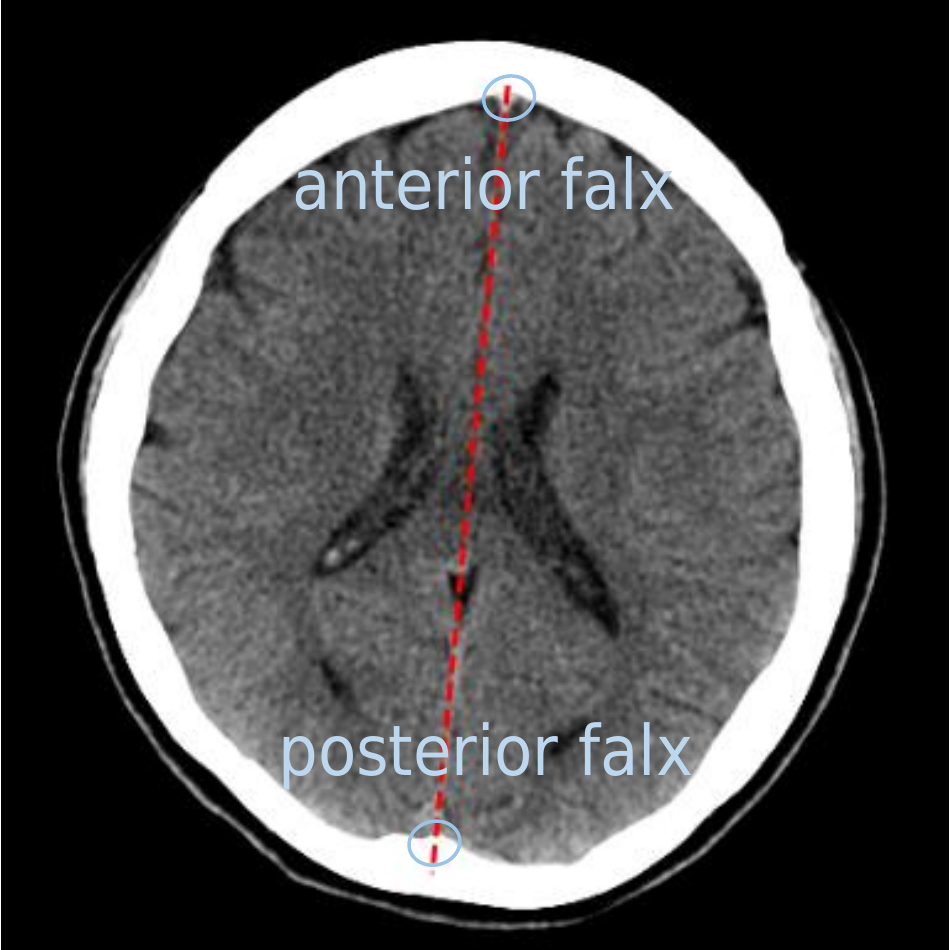}
         \captionsetup{justification=centering}
         \caption{No MLS\\\quad }
         \label{fig:mls_healthy}
     \end{subfigure}
     \hfill
     \begin{subfigure}[b]{0.24\textwidth}
         \centering
         \includegraphics[width=\textwidth, height=\textwidth]{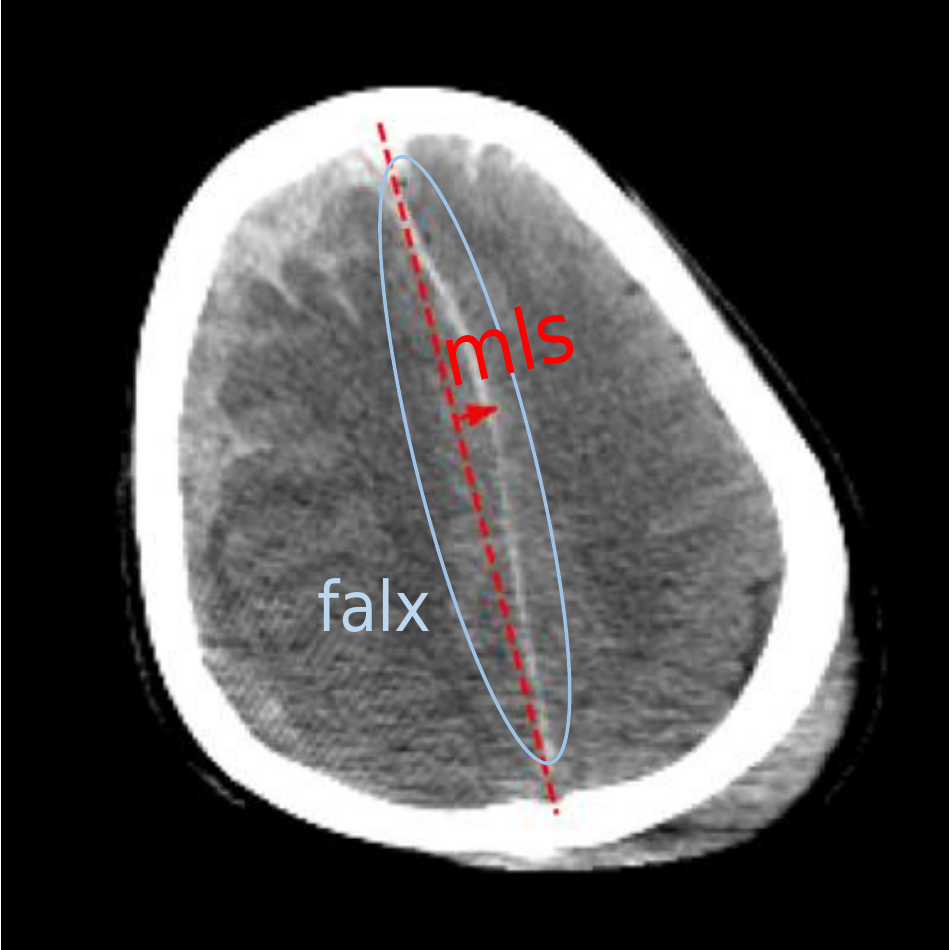}
         \captionsetup{justification=centering}
         \caption{MLS on falx\\ \quad}
         \label{fig:mls_1}
     \end{subfigure}
     \hfill
     \begin{subfigure}[b]{0.24\textwidth}
         \centering
         \includegraphics[width=\textwidth, height=\textwidth]{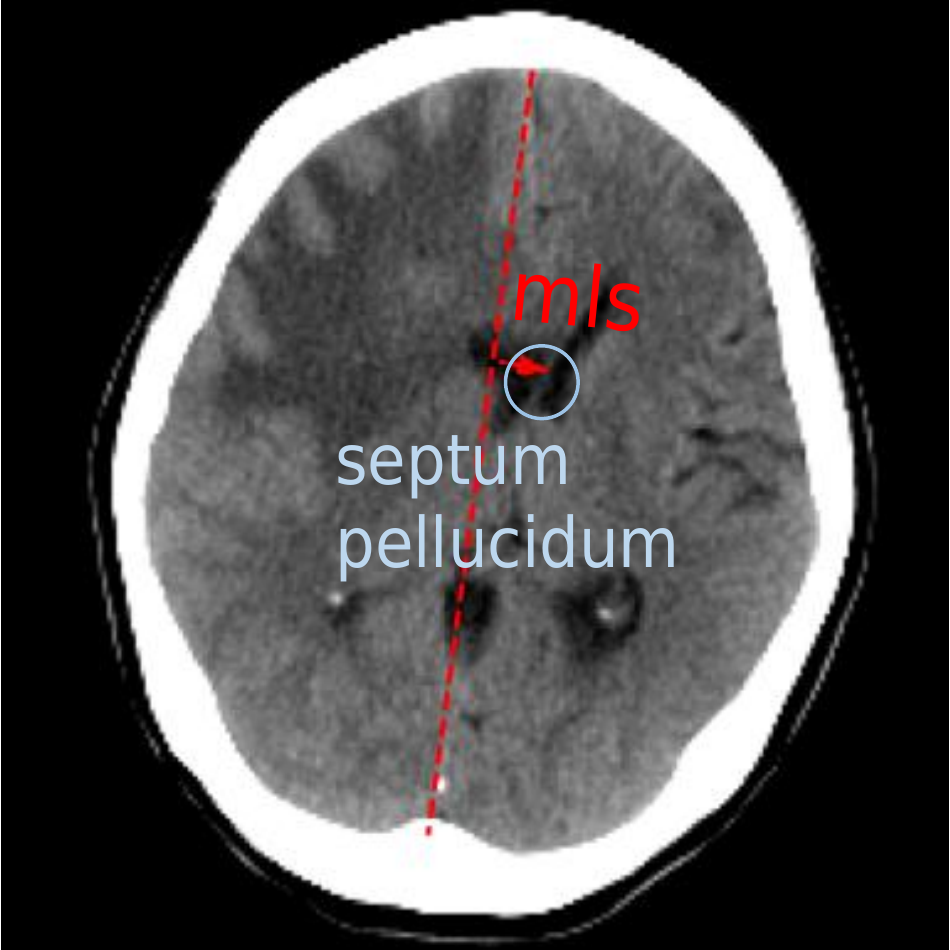}
        \captionsetup{justification=centering}
         \caption{MLS on septum pellucidum}
         \label{fig:mls_2}
     \end{subfigure}
          \hfill
     \begin{subfigure}[b]{0.24\textwidth}
         \centering
         \includegraphics[width=\textwidth, height=\textwidth]{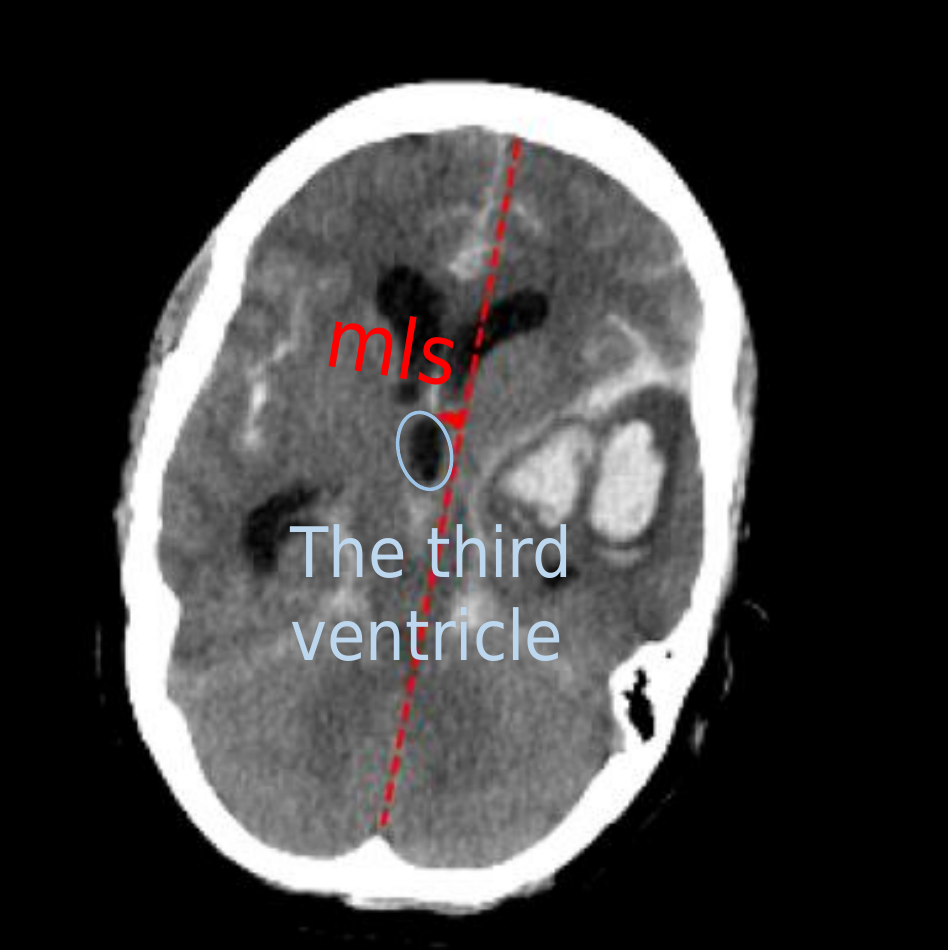}
         \captionsetup{justification=centering}
         \caption{MLS on the third ventricle}
         \label{fig:mls_3}
     \end{subfigure}
        \caption{Examples of head CT scans to illustrate how radiologists measure MLS. Dash red line connecting the anterior falx and posterior falx denotes a hypothetical normal midline. Blue circles denote the shifted landmarks. Perpendicular red lines from the shifted landmarks to normal midline are measured as MLS scale.}
        \label{fig:mls_example}
\end{figure}

Intracranial hemorrhage (ICH) refers to brain bleeding within the skull, a serious medical emergency that would cause severe disability or even death~\cite{ref_ich0}. 
A characteristic symptom of severe ICH is brain midline shift (MLS), which is the lateral displacement of midline cerebral structures (see Fig.~\ref{fig:mls_example}). 
MLS is an important and quantifiable indicator of the severity of mass effects and the urgency of intervention~\cite{ref_ich1,ref_ich2,ref_mls6}. 
For instance, the 5 millimeters ($mm$) threshold of MLS is frequently used to determine whether immediate intervention and close monitoring is required~\cite{ref_ich3}. 
MLS quantification demands high accuracy and efficiency, which is difficult to achieve with manual quantification, especially in emergencies, due to the variability in shift regions, unclear landmark boundaries, and non-standard scanning pose. 
An automated MLS quantification algorithm that can immediately and accurately quantify MLS is highly desirable to identify urgent patients for timely treatment. 

To measure MLS, clinicians usually first identify a few CT slices with large shifts and then measure and identify the maximum deviation of landmarks such as the septum pellucidum, third ventricle, or falx from their normal counterpart as the final MLS distance (see examples in Fig.~\ref{fig:mls_example}). 
Such a clinical fashion of MLS quantification can be difficult to be translated into a well-defined automation process. 
Currently, there are only limited studies on automated MLS quantification, using different strategies and varied labeling requirements. 
Nguyen et al. proposed a landmark-based method that relies on anatomical markers to determine the location of the deformed midline~\cite{ref_mls6}. However, this method can only apply to cases where MLS appears on these specific marker regions. 
Liao et al. adopted a symmetric-based method to seek a curve connecting all deformed structures~\cite{ref_mls7}, which is difficult to generalize due to over-simplified anatomical assumptions and sensitivity to patients' scan poses.
A few recent works try to overcome these limitations by using stronger supervision with dense labeling. 
Some studies formulated MLS quantification as a midline segmentation task~\cite{ref_mls1,ref_mls2,ref_mls3}, by delineating the intact midline as labels to supervise the training of segmentation models.
Another study designed a hemisphere segmentation task to quantify MLS~\cite{ref_mls5}, which requires pixel-wise annotation for each slice.
However, obtaining such dense annotations is very costly and time-consuming, while may not be necessary for MLS quantification.

To tackle these limitations, we propose to fit MLS quantification into a deformation prediction problem by using semi-supervised learning (SSL) with only limited annotations. Our framework avoids the strong dependency on specific landmarks or over-simplified assumptions in previous methods while not increasing the labeling efforts. 
We aim to use only sparse and weak labels as ground truth supervisions, which are just one shifted landmark and its normal counterpart on a limited number of slices provided by radiologists, but we try to fully exploit the unlabeled slices and non-MLS data to impose extra regularization for the sparse-to-dense extension.
Existing SSL methods typically use a partially trained model with labeled data to generate pseudo labels for unlabeled data, assuming that labeled and unlabeled data are generally similar. 
These methods can be sub-optimal in our case as labeled slices of MLS usually present the largest deformation while unlabeled slices contain only minor or no deformation.
Instead, we propose our SSL strategy by generating a corresponding non-MLS image for each unlabeled MLS slice with generative models and regularizing that the deformation field should warp the generated non-MLS images into the original MLS ones. 
However, as we only have volume-wise labels for MLS and non-MLS classification, it can be difficult to train a slice-wise discriminator as required by many generative models such as GANs~\cite{ref_gan}. Fortunately, the recently proposed diffusion models~\cite{ref_diffusion3}, which prove to have strong power in both distribution learning and image generation without dependency on discriminators, can be a potentially good solution.

In this work, we propose a novel semi-supervised learning framework based on diffusion models to quantify the brain MLS from head CT images with deformation prediction. 
Our method effectively exploits supervision and regularization from all types of available data including MLS images with sparse ground truth labels, MLS images without labels, and non-MLS images.
We validate our method on a real clinical head CT dataset, showing effectiveness of each proposed component. Our contributions include: (1) innovating an effective deformation strategy for brain MLS quantification, (2) incorporating diffusion models as a representation learner to extract features reflecting where and how an MLS image differs from a non-MLS image, and (3) proposing a diffusion model-based semi-supervised framework that can effectively leverage massive unlabelled data to improve the model performance.

\section{Methods}

\begin{figure}[t]
\includegraphics[width=\textwidth]{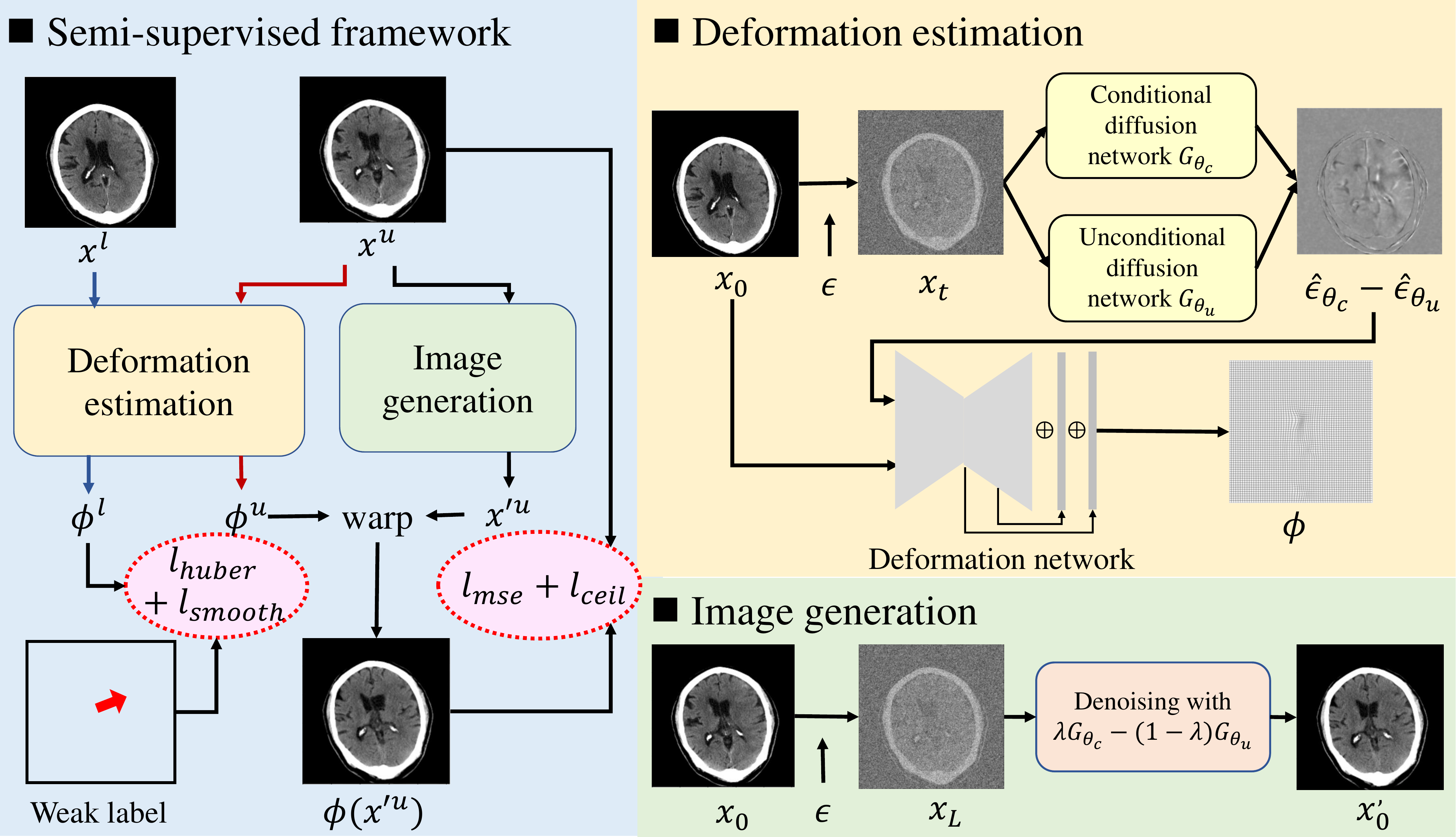}
\caption{The pipeline of our proposed semi-supervised deformation strategy for MLS quantification. The labeled image $x^{l}$ is supervised by sparse labels and the unlabeled image $x^{u}$ is self-supervised with generated negative image.}
\label{framework}
\end{figure}

Fig.~\ref{framework} illustrates our diffusion model-based semi-supervised learning framework for MLS quantification via deformation prediction. In Sec.~\ref{method1}, we introduce our deformation prediction by using only sparse supervision. In Sec.~\ref{method2}, we propose to incorporate non-MLS data for representation learning. In Sec.~\ref{method3}, we describe how to utilize unlabeled MLS images for sparse-to-dense regularization.

\subsection{MLS Quantification through Deformation Estimation}
\label{method1}
Our proposed deformation strategy for brain MLS quantification aims to find an optimal deformation field $\phi$ so that an MLS image can be regarded as a hypothetically non-MLS image warped with this deformation field. The deformation field can be parameterized by a function with high complexity so that it does not explicitly rely on a single landmark or over-simplified symmetric assumptions, which naturally overcomes the limitations of existing methods. We apply a learning-based framework to parameterize the deformation field with a U-Net shape neural network. The output of the network is the stationary velocity field $v$. The diffeomorphic deformation field $\phi$ is then calculated through the integration of the velocity field, similarly to VoxelMorph~\cite{ref_vox} for image registration. The learning process is supervised by sparse deformation ground truth. For each labeled slice, we have the ground truth $\mathbf{y}=(y_1, y_2)$, which is a two-dimensional vector directing from shifted landmark point toward its presumably normal location (the red arrow in Fig.~\ref{framework}). The predicted deformation $\hat{\mathbf{y}}$ is bilinearly interpolated at the shifted landmark point from the deformation field, which is also a two-dimensional vector. To alleviate the influence of a few extremely large deformation points and increase model's robustness, we use Huber loss to measure the similarity between the predicted deformation and the label:
\begin{equation}
\label{eq6}
l_{\text{huber}}(y_d, \hat{y}_d)=\left\{
\begin{aligned}
&|y_d-\hat{y}_d|, &|y_d-\hat{y}_d| \geq c, \\
&\frac{(y_d-\hat{y}_d)^2+c^2}{2c}, &|y_d-\hat{y}_d| < c.
\end{aligned}
\right.
\end{equation}
where $d \in \{1,2\}$.
The hyperparameter $c$ defines the range for absolute error or squared error. 
We also encourage a smooth deformation field with a diffusion regularizer on the spatial gradients of deformation $\phi$ to avoid a discontinuous deformation field:
\begin{equation}
l_{\text{smooth}}=\sum_j\sum_k\Vert\phi_{jk}-\phi_{(j-1)k}\Vert^2+\Vert\phi_{jk}-\phi_{j(k-1)}\Vert^2,
\end{equation}
As the deformation can be extremely large in our case, and meanwhile to force a smooth transition between the deformation peak and its adjacent pixels, we use a coarse-to-fine manner, where velocity fields are generated through upsampling with skip connection to progressively aggregate features of different scales.

\subsection{Learning Negative Patterns from Non-MLS Images}
\label{method2}
In order to learn a deformation field to warp a non-MLS image into MLS one, ideally we would need a pair of non-MLS and MLS images for network training, which however does not exist in practice. 
Lacking such information makes the network difficult to learn.
A naive solution is to generate a corresponding non-MLS image. However, generated images entail some randomness and can often lack important details. Depending too much on such fake inputs can lead to poor robustness.
Inspired by the score-matching interpretation of diffusion models~\cite{ref_diffusion5}, we propose to learn the non-MLS distribution from massive amount of negative cases. Given an MLS image, we can evaluate which parts of the image make it different from a non-MLS image. This deviation can serve as latent features that help the deformation network with deformation prediction.

Diffusion models, especially DDPM~\cite{ref_diffusion2}, define a forward diffusion process as the Markov process progressively adding random Gaussian noise to a given image and then trying to approximate the reverse process by a Gaussian distribution. The forward process can be simplified by a one-step sampling:
$x_t = \sqrt{\alpha_t}x_0+\sqrt{1-\alpha_t}\epsilon$,
where $\alpha_t:=\prod^t_{s=0}1-\beta_t$, and $\beta_t$ are predefined variance schedule. $\epsilon$ is sampled from $\mathcal{N}(0,I)$. The mean $\mu_\theta(x_t,t)$ and variance $\Sigma_\theta(x_t,t)$ of the reverse process can be parameterized by neural networks. A popular choice is to re-parameterize $\mu_\theta(x_t,t)$ so that $\hat{\epsilon}_\theta(x_t, t)$ instead of $\mu_\theta(x_t,t)$ is estimated by neural networks to approximate the noise $\epsilon$. Moreover, the output of the diffusion network $\epsilon(x_t, t)$ is actually a scaled score function $\nabla \log p(x_t)$ as it moves the corrupted image towards the opposite direction of the corruption.~\cite{ref_diffusion6}. 

As a result, through pre-training one unconditional diffusion model trained with all data (denoted as $\mathcal{U}$) and one conditional diffusion model trained with only non-MLS data (denoted as $\mathcal{C}$), the subtraction of two outputs 
\begin{equation}
\hat{\epsilon}_{\theta_\mathcal{U}}(x_t,t)-\hat{\epsilon}_{\theta_\mathcal{C}}(x_t,t) \propto \nabla \log p(x_t|n) - \nabla \log p(x_t) =  \nabla \log p(n|x_t),
\end{equation}
can be regarded as the gradient of class prediction ($n=1$ for non-MLS and 0 otherwise) w.r.t to the input image, which reflects how the input images deviate from a non-MLS image. This latent contains information regarding how to transform the MLS positive image into a non-MLS one and therefore is helpful for training the deformation network. Moreover, this feature representation exhibits less fluctuation toward the randomness of the additive noise because the stochastic parts are eliminated through subtraction. It is more stable than the predicted noise or generated MSL negative images. For training, we randomly sample $t$ from $0$ to the diffusion steps $T_{\text{train}}$, while for inference we fix it to be a certain value. We examine the effects of this value in Section \ref{ablation1}.

\subsection{Semi-Supervised Deformation Regularization}
\label{method3}
Deformation estimation is a dense prediction problem, while we only have sparse supervision. This can lead to flickering and poor generalizability if the deformation lacks certain regularization. On the other hand, we have a significant amount of unlabeled data from the MLS volumes that is potentially helpful. Therefore, we propose to include these unlabeled data during training in a semi-supervised manner, so that unlabeled data can provide extra regularization for training or produce additional training examples based on noisy pseudo labels. Many existing semi-supervised methods seek to use the prediction for unlabeled data given by the same or a twin network as pseudo-labels and then supervise the model or impose some regularization with these pseudo-labels. However, these methods hold a strong assumption that labeled and unlabeled data are drawn from the same distribution, which is not true in our case because most labeled data are with large deformation while unlabeled data are with minor or no deformation. Therefore, we want to find another type of pseudo-label to bypass the distribution assumption. As the deformation field is assumed to warp a hypothetically normal image into an MLS one, we generate hypothetically non-MLS images $x'_0$ using pre-trained diffusion models through classifier-free guidance~\cite{ref_diffusion4}:
\begin{equation}
\hat{\epsilon}(x_t, t) = \gamma \hat{\epsilon}_{\theta_\mathcal{C}}(x_t,t) + (1-\gamma)\hat{\epsilon}_{\theta_\mathcal{U}}(x_t,t),
\end{equation}
where $\gamma$ is a hyper-parameter controlling the strength of the condition. We compare $x'_0$  warped with the deformation field $\phi(x'_0)$ and calculate its similarity with the original $x_0$ through MSE loss. As it can be difficult for the generated image to be fully faithful to the original image because the generative process entails a lot of random sampling, this $l_\text{mse}$ can only serve as noisy supervision. Therefore, instead of generating $x'_0$ ahead of deformation network training, we generate it in an ad-hoc way so that the noisy effects can be counteracted. 

The final MLS measurement is estimated by calculating the length of the maximum displacement vector from the predicted deformation field, so it is more sensitive to over-estimation.  And our results also show most of the errors come from over-estimation. As for unlabelled slices, we still have the prior that its MLS cannot be larger than the MLS of that specific volume $\delta$, we propose to incorporate an additional ceiling loss to punish the over-estimation:
\begin{equation}
l_{\text{ceil}} = \sum_{j}\sum_{k} \max(0, || \phi_{jk} ||-\delta). 
\end{equation}
Overall, the loss term is a combination of supervised loss and unsupervised loss, with a weight term controlling the relative importance of each loss term:
\begin{equation}
l = l_{\text{huber}} +w_1 l_{\text{smooth}} + u(i)(l_\text{mse} + w_2 l_{\text{ceil}}),
\end{equation}
where $w_1$ and $w_2$ are two fixed weight terms and  $u(i)$ is a time-varying weight term that is expected to gradually increase as the training iteration $i$ progresses so that the training can converge quickly through strong supervision first and then refine and enhance generalizability via unsupervised loss.

\section{Experiments and Results}
\subsection{Data Acquisition and Preprocessing}
We retrospectively collected anonymous thick-slice, non-contrast head CT of patients who were admitted with head trauma or stroke symptoms and diagnosed with various subtypes of intracranial hemorrhage, including epidural hemorrhage, subdural hemorrhage, subarachnoid hemorrhage, intraventricular hemorrhage, and intraparenchymal hemorrhage, between July 2019 and December 2019 in the Prince of Wales Hospital, a public hospital under the Hospital Authority of Hong Kong. The ethics approval was obtained from the Joint Chinese University of Hong Kong-New Territories East Cluster Clinical Research ethics committee.
The eligible patients comprised 2793 CT volumes, among them 124 are MLS positive cases. The MLS ranges between $2.24mm$ and $20.12mm$, with mean value of $8.34mm$ and medium value of $8.73mm$. 
The annotation was performed by two trained physicians and verified by one experienced radiologist (with over 10 years of clinical experience on ICH). The labeling process followed a real clinical measurement pipeline, where the shifted landmark, anterior falx point, and posterior falx point were pointed out, and the length of the vertical line from the landmark to the line connecting the anterior falx point and the posterior falx point was the measured MLS value. For each volume, a few slices with large deformation were separately measured and annotated while the shift of the largest one served as the case-level label. On average, 4 out of 30 slices of each volume were labeled. We discarded the first 8 and the last 5 slices as they are mainly structures irrelevant to MLS. For pre-processing, we adjusted the pixel size of all images to $0.86mm$ and then cropped or padded the resulting images to the resolution of 256 $\times$ 256 pixels. The HU window was set to 0 and 80. We applied intensity clipping (0.5 and 99.5 percentiles) and min-max normalization (between -1 and 1) to each image. Random rotation between $-15^{\circ}$ and $15^{\circ}$ was used for data augmentation.

\subsection{Implementation Details}
For the diffusion network, we use the network architecture designed in DDPM~\cite{ref_diffusion3} and set the noise level from $10^{-4}$ to $2 \times 10^{-2}$ by linearly scheduling with $T_{\text{train}}=1000$. For non-MLS image generation, we apply the Denoising Diffusion Implicit Model (DDIM)~\cite{ref_diffusion1} with 50 steps and set the noise scale to 15 to shorten the generative time. We set the hyper-parameters as $\alpha=1$, $\beta=1$, $c=3$ and $\gamma=2$. $u(i)$ is set from 1 to 10 with the linear schedule. The diffusion models are trained by the AdamW optimizer with an initial learning rate of $1\times10^{-4}$, batch size 4, for $2\times10^5$ iterations. We up-sample the MLS positive data by $10\times$ when training the unconditional diffusion model. The deformation network is trained by the AdamW optimizer with an initial learning rate of $1\times 10^{-4}$, batch size 16, for 100 epochs. All models are implemented with PyTorch 1.12.1 using one Nvidia GeForce RTX 3090 GPU.

\begin{table*}[t!]
\centering
\renewcommand\arraystretch{1.2}
\caption{Comparison of different methods with 5-fold cross-validation.}
\begin{center}
    \resizebox{1.0\textwidth}{!}{%
        \setlength\tabcolsep{4.0pt}
        \begin{tabular}{l|cc|cc|cc}
        \toprule[1.5pt]
        
        \multirow{2}{*}{Methods} &\multicolumn{2}{c|}{Training data}&\multicolumn{2}{c|}{Volume-wise}  & \multicolumn{2}{c}{Slice-wise} \\
        \cline{2-7}

        {} &Labeled&Unlabeled& \makecell[c]{MAE$\downarrow$\\(mm)} & \makecell[c]{RMSE$\downarrow$\\(mm)} & \makecell[c]{MAE$\downarrow$\\(mm)} & \makecell[c]{RMSE$\downarrow$\\(mm)}  \\

        \hline
        \hline

        Regression &\checkmark&	&3.91 &4.90 &3.56 &4.16\\ 

        Deformation &\checkmark&	& 	3.80 & 4.47 &2.51&3.17\\

        \hline
        Mean-Teacher~\cite{ref_semi1} &\checkmark&	\checkmark&	2.89 &  3.67 & 2.43&3.22\\ 
        
        CPS~\cite{ref_semi2} &\checkmark&	\checkmark& 2.72 &3.42 &2.38&3.15\\ 
        
        Ours &\checkmark&	\checkmark&  \textbf{2.43}  &  \textbf{3.17} &  \textbf{2.25}&\textbf{3.09}\\
        \bottomrule[1.5pt]
        
        \end{tabular}}
\end{center}
\label{table_1}
\end{table*}





\subsection{Quantification Accuracy and Deformation Quality}
We evaluate the performance of our quantification strategy through mean absolute error (MAE) and root mean square error (RMSE). For volume-wise evaluation, we measure the maximum deformation of each slice of the whole volume and select the largest one as the final result. We also report the slice-wise evaluation, which is calculated based on labeled slices. This error can reflect how the models perform on slices with relatively large deformation. 
Since existing MLS estimation methods require different types of labels from ours, it is difficult to directly compare with those methods. We therefore first compare our deformation-based strategy with a regression-based strategy, which uses DenseNet-121~\cite{ref_dense} to directly predict the slice-wise MLS. We also compare our proposed semi-supervised learning approach with two popular semi-supervised learning methods, that are Mean-Teacher~\cite{ref_semi1} and Cross Pseudo Supervision (CPS)~\cite{ref_semi2}, which are implemented into our deformation framework. The results are given in Table~\ref{table_1}, which are based on 5-fold cross-validations. 

From the results, we can see that when only using labeled MLS slices for model learning, our deformation strategy already shows better performance than the regression model. This may attribute to that our deformation model learns the knowledge of both MLS values and locations while a regression model only captures the MLS value information. 
This difference can be further enlarged if we consider slice-wise performance. 
Moreover, all three semi-supervised learning methods, i.e., Mean-Teacher, CPS, and ours, consistently improve the performance of deformation prediction, showing the benefits and importance of incorporating unlabeled data into model learning.
Our semi-supervised learning method based on diffusion models achieves better quantification results than Mean-Teacher and CPS, significantly reducing the volume-wise MAE from 3.80$mm$ to 2.43$mm$. 
An interesting observation is that the unlabeled data contribute more to the volume-wise evaluation than the slice-wise evaluation. By inspecting the prediction, we find that the deformation prediction trained with labeled data tends to overestimate the deformation of slices with little or no deformation, which makes the volume-wise prediction error-prone. As most unlabeled data are slices with minor shifts, incorporating these data for semi-supervised learning can impose constraints to avoid large deformation, which greatly improves the model's robustness.


\begin{figure}[t!]
\begin{minipage}[b]{\linewidth}
     \centering
          \begin{subfigure}[b]{0.24\textwidth}
         \centering
         \includegraphics[width=\textwidth, height=\textwidth]{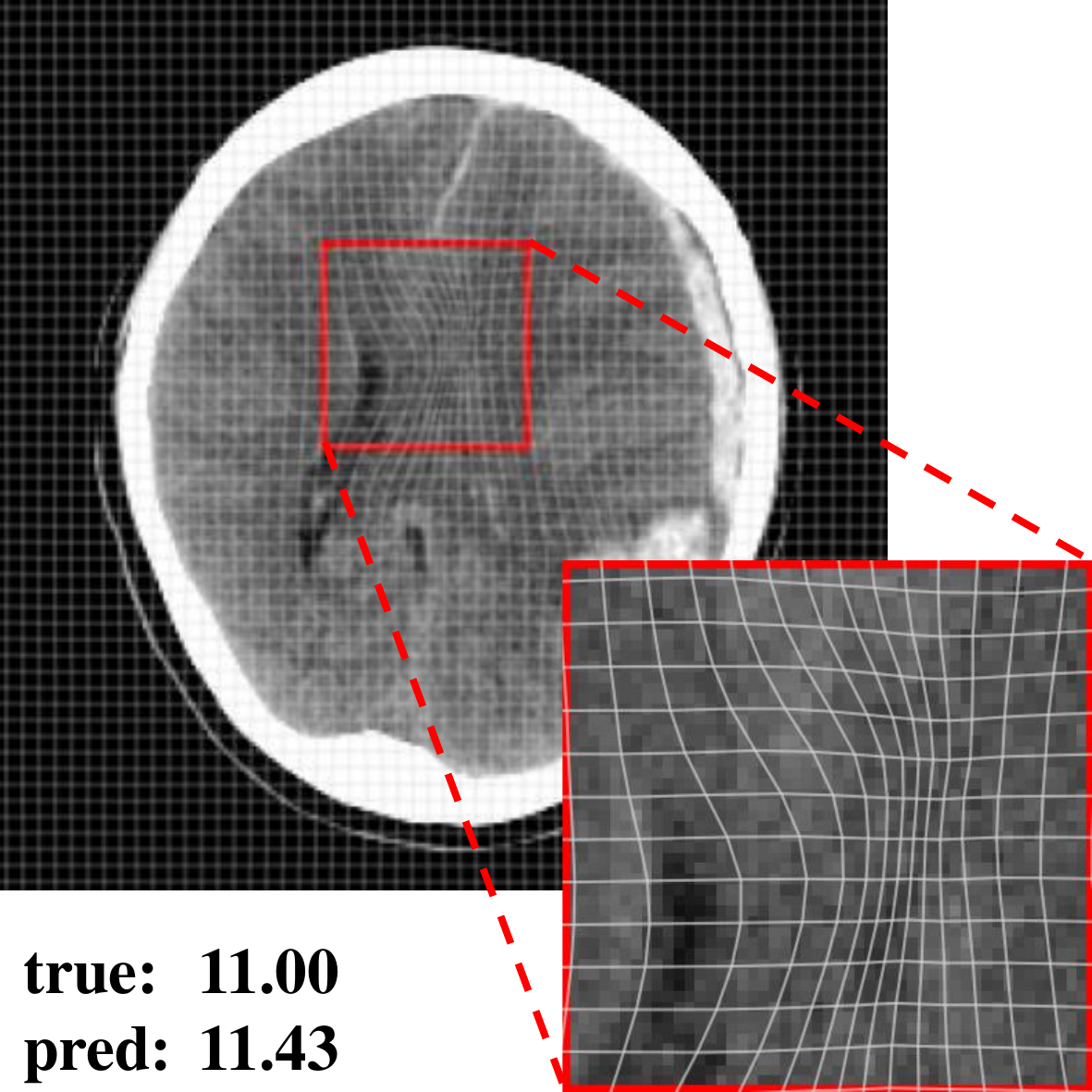}
         \captionsetup{justification=centering}
     \end{subfigure}
     \hfill
     \begin{subfigure}[b]{0.24\textwidth}
         \centering
         \includegraphics[width=\textwidth, height=\textwidth]{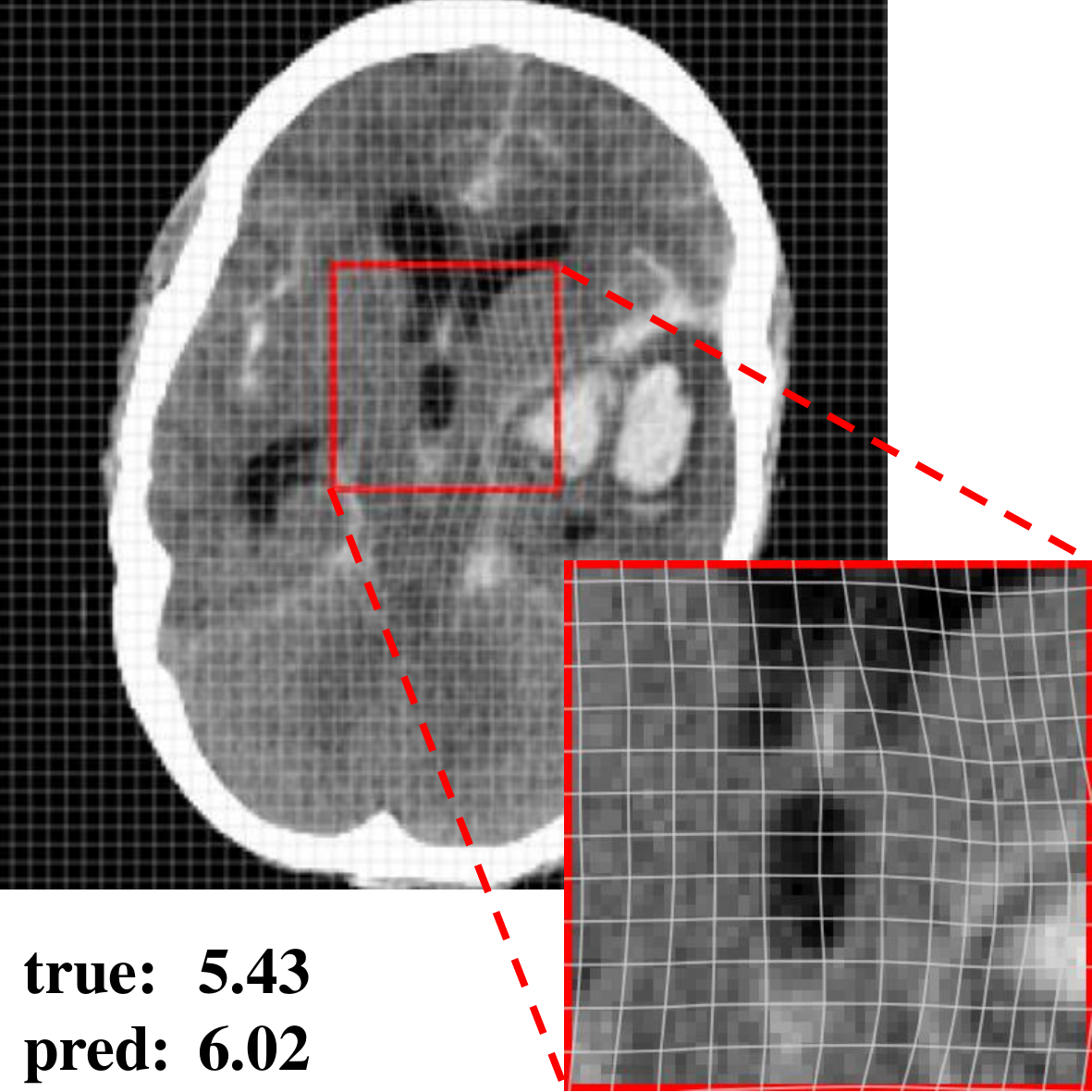}
         \captionsetup{justification=centering}
     \end{subfigure}
     \hfill
     \begin{subfigure}[b]{0.24\textwidth}
         \centering
         \includegraphics[width=\textwidth, height=\textwidth]{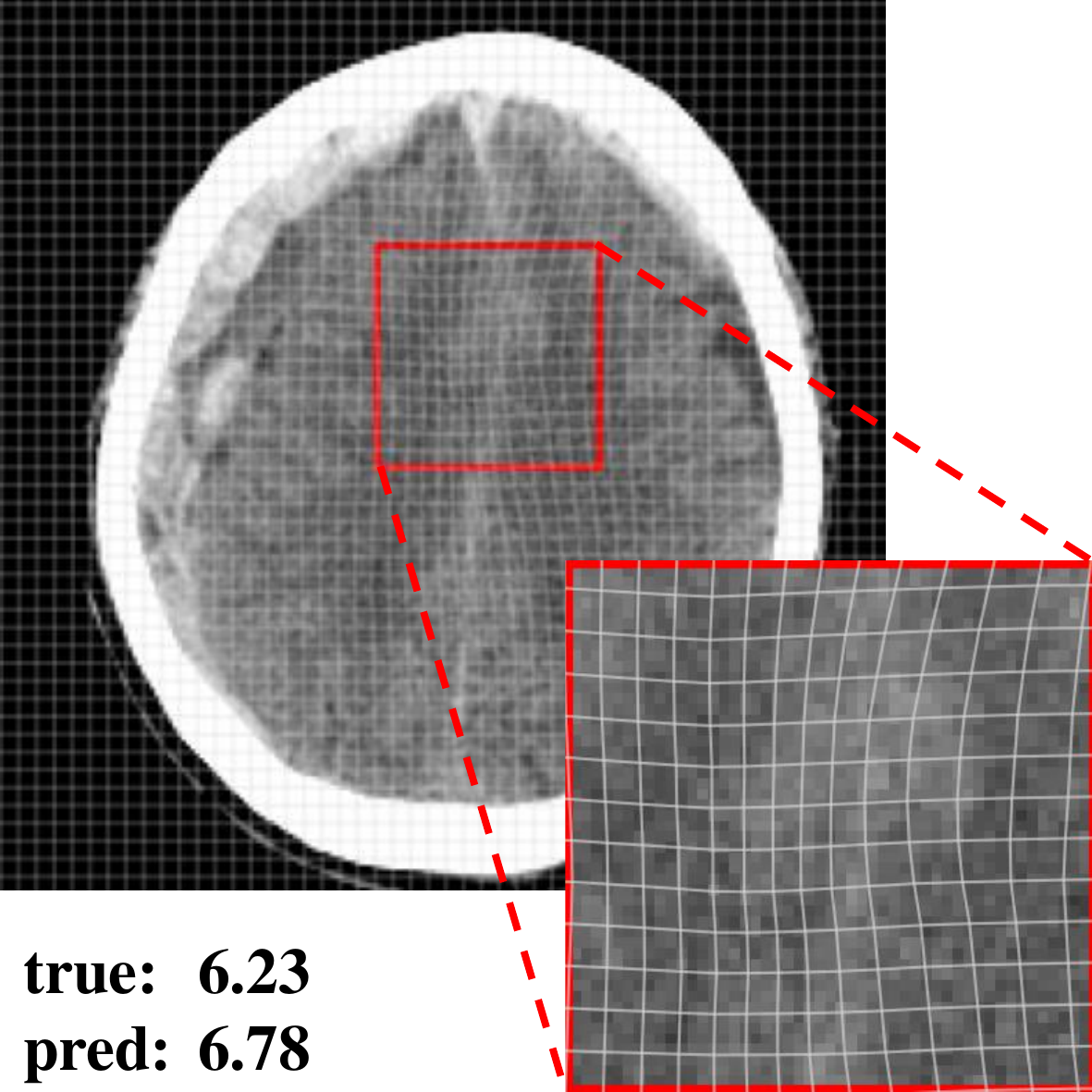}
         \captionsetup{justification=centering}
     \end{subfigure}
          \hfill
     \begin{subfigure}[b]{0.24\textwidth}
         \centering
         \includegraphics[width=\textwidth, height=\textwidth]{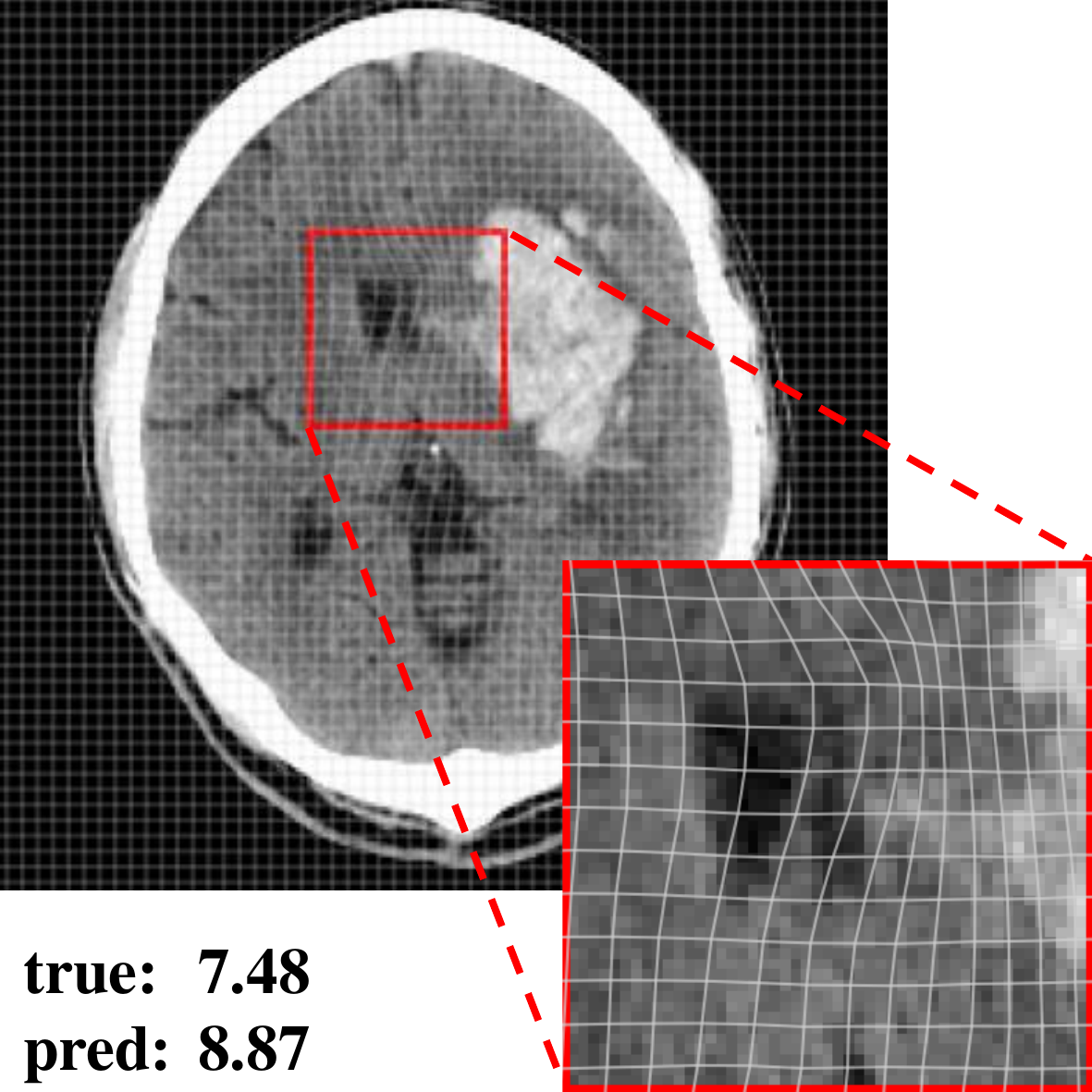}
         \captionsetup{justification=centering}
     \end{subfigure}
     \vspace {-0.3cm} 
     \caption*{(a) Examples of predicted deformation on MLS images}
      \vspace {0.5cm} 
     \label{fig:deformation1}
    \end{minipage}
   
    \begin{minipage}[b]{\linewidth}
    \begin{subfigure}[b]{0.24\textwidth}
         \centering
         \includegraphics[width=\textwidth, height=\textwidth]{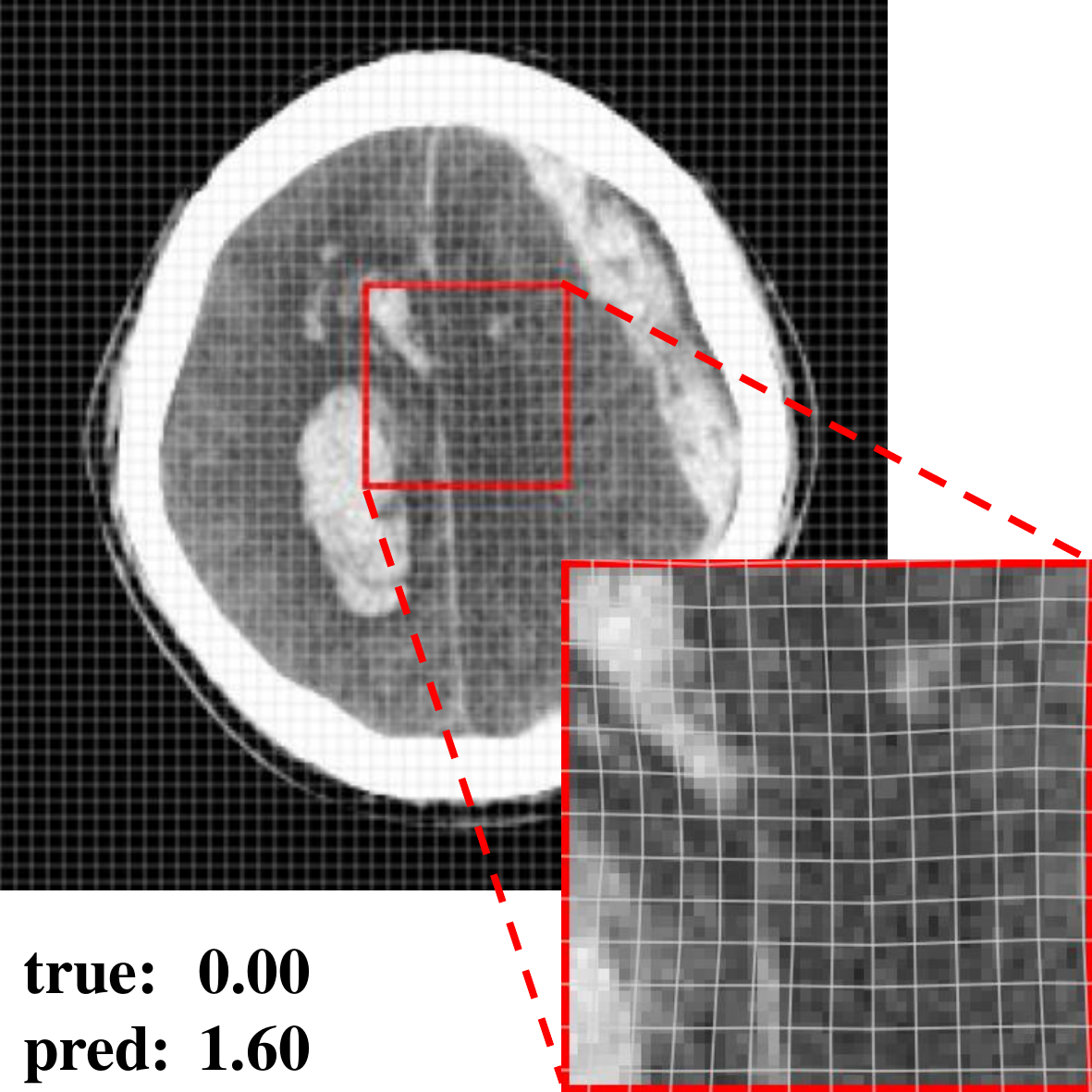}
         \captionsetup{justification=centering}
     \end{subfigure}
     \hfill
     \begin{subfigure}[b]{0.24\textwidth}
         \centering
         \includegraphics[width=\textwidth, height=\textwidth]{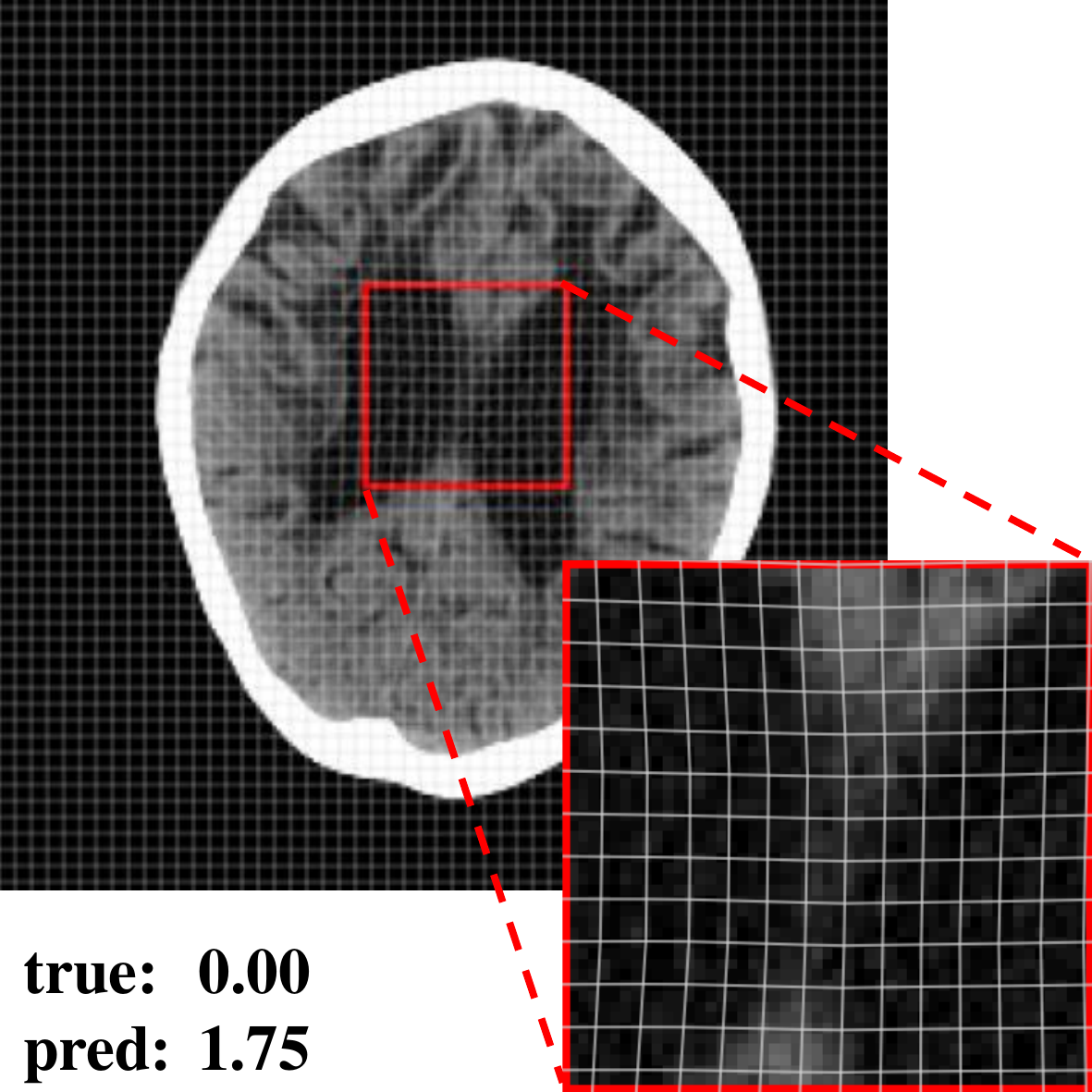}
         \captionsetup{justification=centering}
     \end{subfigure}
     \hfill
     \begin{subfigure}[b]{0.24\textwidth}
         \centering
         \includegraphics[width=\textwidth, height=\textwidth]{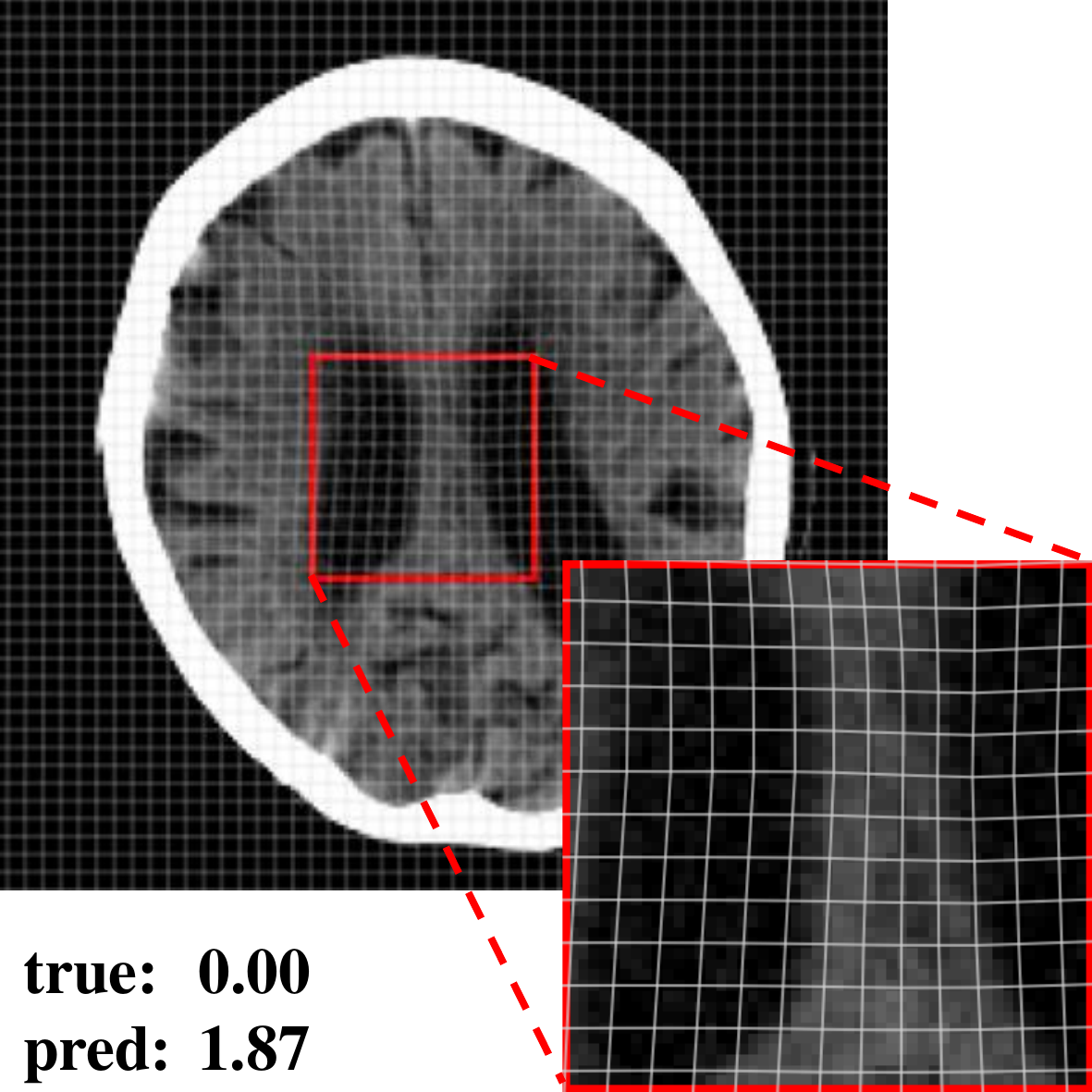}
         \captionsetup{justification=centering}
     \end{subfigure}
          \hfill
     \begin{subfigure}[b]{0.24\textwidth}
         \centering
         \includegraphics[width=\textwidth, height=\textwidth]{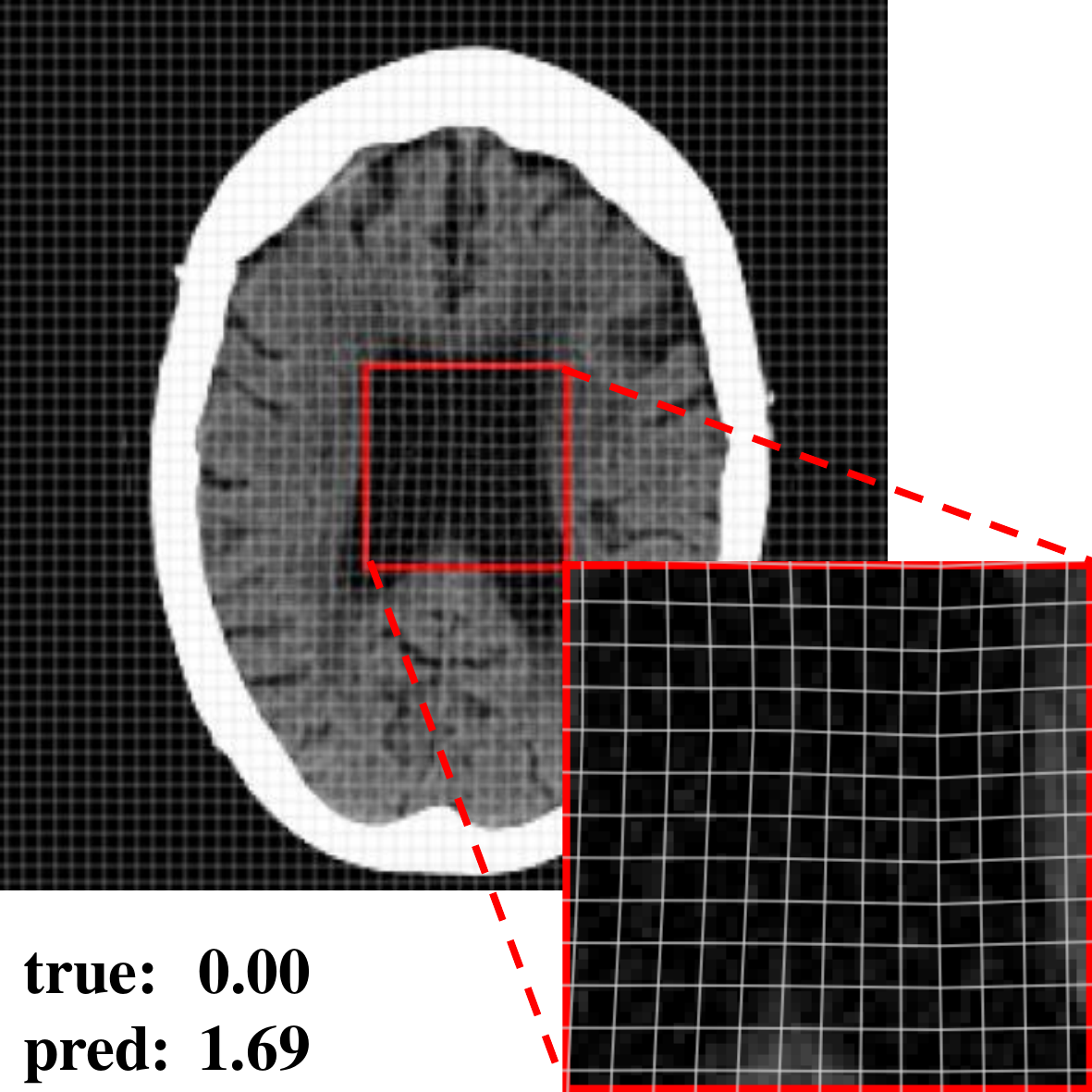}
         \captionsetup{justification=centering}
     \end{subfigure}
     \vspace {-0.3cm} 
        \caption*{(b) Examples of predicted deformation on non-MLS images}
        \label{fig:deformation2}
        \end{minipage}
        \caption{Predicted deformation on (a) MLS images. (b) non-MLS images. The regions with the largest deformation are highlighted. Slice-wise predicted MSL and ground truth are provided.}
        \label{fig:deformation3}
\end{figure}
We also visualize the predicted deformation field of several sample cases. From Fig.~\ref{fig:deformation3} (a), we can see the model can well posit the location where the maximum shift appears and push it to its hypothetically normal counterpart. The largest deformation happens exactly at the site with the maximum shift. To validate the robustness of our model, we also select several patients diagnosed with no MLS and plot the predicted deformation of these samples. As can be seen in Fig.~\ref{fig:deformation3} (b), our method is able to provide a reasonable prediction for non-MLS images by outputting much smaller values than that for MLS images.
Our model's predictions for non-MLS images are not exactly zero are caused on one hand by that even for a completely healthy person, the midline cannot be perfectly aligned due to multiple factors such as scan pose, on the other hand, our models tend to overestimate the shift because we are calculating the maximum deformation as final measurement. 

\subsection{Ablation Study}
We conduct several ablation experiments to study the effects of several components in our proposed framework on the model performance. The volume-wise results reported are trained on four folders and tested on one folder.
\vspace{-5mm}
\subsubsection{Effects for representation learning.}
\label{ablation1}
\begin{figure}[t!]
\begin{minipage}[b]{.5\linewidth}
\centering
\begin{tabular}{cc|cc}
\toprule[1.5pt]
\multicolumn{2}{l|}{Methods} & \makecell[c]{MAE$\downarrow$\\(mm)} & \makecell[c]{RMSE$\downarrow$\\(mm)} \\
\hline
\hline
\multicolumn{2}{l|}{Fully-supervised} &  3.61 &   4.47 \\

\multicolumn{2}{l|}{ + Representation} & 	3.22 & 3.69 \\ 
\hline
\multicolumn{2}{l|}{Semi-supervised} &	2.61 &  3.24 \\ 

\multicolumn{2}{l|}{+ Representation}& 2.45 &3.05 \\ 
\bottomrule[1.5pt]
\end{tabular}
\captionof{table}{Effects of the representation.}
\label{latent_res}
\end{minipage}
\begin{minipage}[b]{.5\linewidth}
\centering
\includegraphics[width=\textwidth, height=0.5\textwidth]{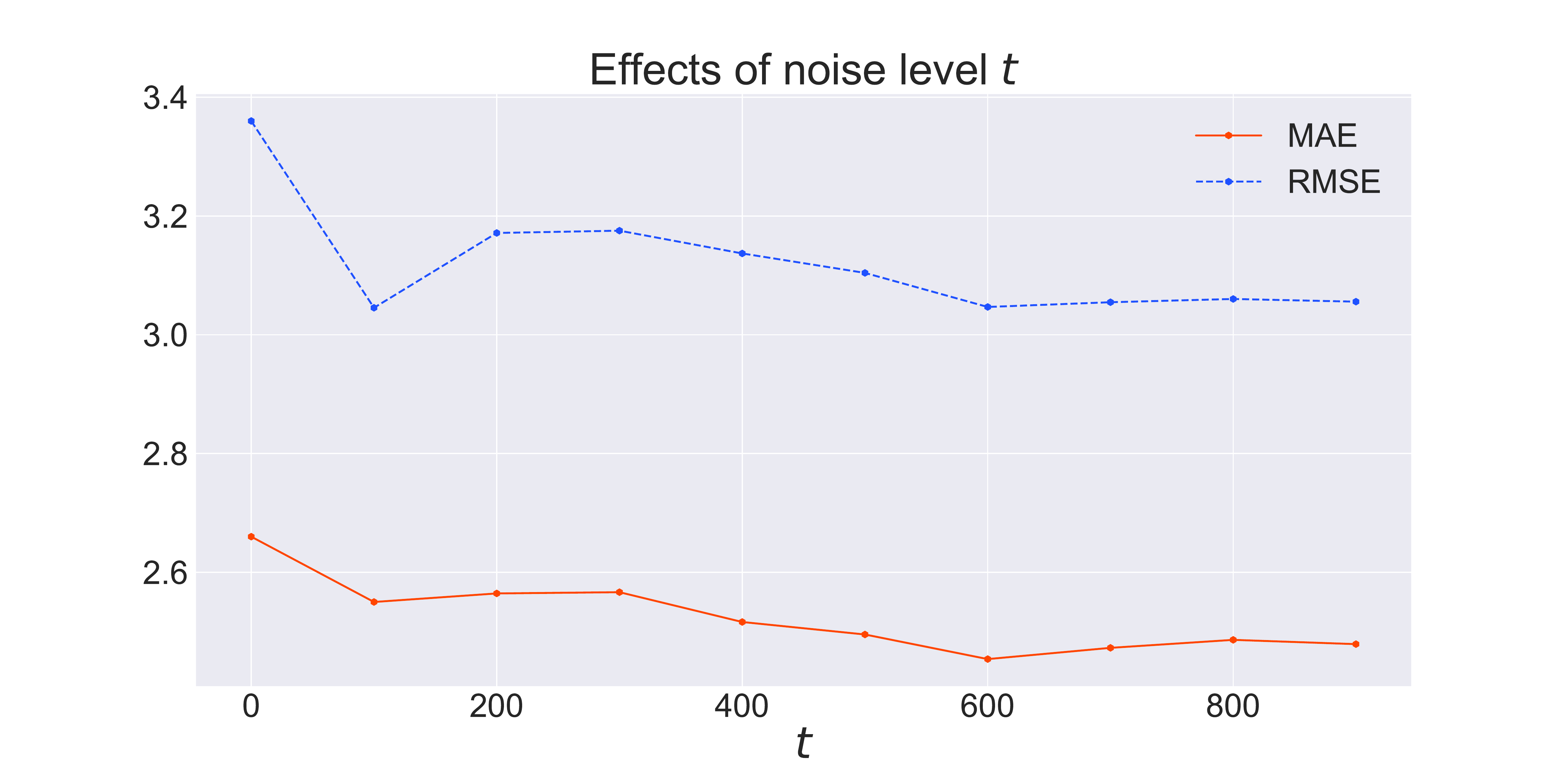}
\caption{Effects of the noise level.}
\label{fig:latent}
\end{minipage}
\end{figure}

We first conduct ablation studies to verify that the latent feature extracted from the two diffusion models is truly useful for deformation prediction. To this end, we select two deformation models, one trained with only labeled data and the other using semi-supervised learning, and compare their performance with and without the extracted representation as input. The results are given in Table~\ref{latent_res}. As expected, incorporating the representation can improve the model performance in both cases.

The noise level is an important component of diffusion models. Only with a proper noise level, can the model accurately estimate the deviation of the image toward the negative sample space. Therefore, we do inference with multiple noise levels and compare its effect on model performance. The results are shown in Fig.~\ref{fig:latent}. Our model is very robust towards this hyper-parameter. As long as $t$ is not too small, the model gives very similar performances. The best performance appears in the middle when $t=600$. This is reasonable as small noise fails to corrupt the original image thus degenerating the performance of score estimation while large noise may obscure too many details of the original image.  

\vspace{-3mm}
\subsubsection{Quantity of unlabeled images.}
To verify the usefulness of unlabeled images, we conduct ablation studies on the number of unlabeled images used. For each experiment, we randomly sample 20\%, 40\%, 60\%, and 80\% volumes, and we incorporate unlabeled slices of these volumes for semi-supervised training. For the rest volumes, we are only using the labeled slices. We also do one experiment that completely removes the uses of unlabeled images. For each experiment, the pre-trained diffusion models are the same, which uses all the data. In other words, these unlabeled images somehow still contribute to the model training. The results are shown in Fig.~\ref{fig:hist} (a). As can be seen, the model performance and robustness can be enhanced as we incorporate more unlabeled images. This provides strong evidence for our claim that our model truly learns valuable information from unlabeled data.
\vspace{-3mm}
\subsubsection{Quantity of non-MLS images.}
\begin{figure}[t]
     \centering
    \includegraphics[width=\textwidth, height=0.3\textwidth]{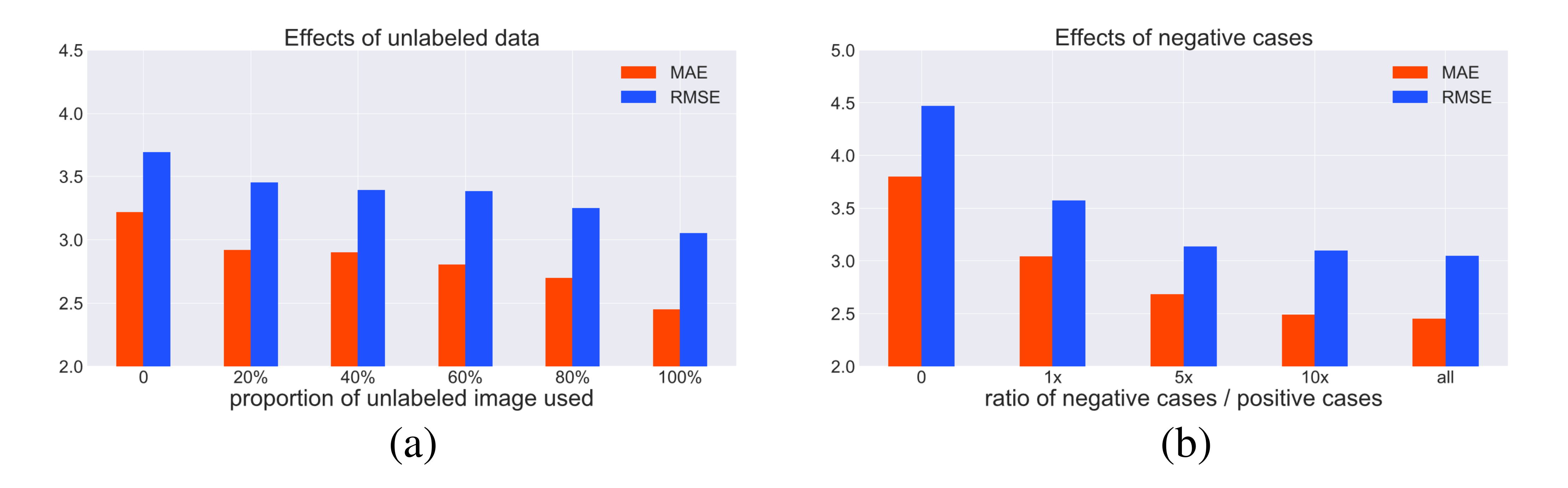}
    \caption{Results of our ablation experiments in terms of: (a) proportion of unlabeled data used, and (b) proportion of negative data used.}
    \label{fig:hist}
\end{figure}

To further measure the benefits of including non-MLS cases, we conduct another ablation study on the proportion of non-MLS cases. Non-MLS cases are used to train diffusion models. As currently, the amount of non-MLS cases is much higher than MLS cases, we upsample the MLS cases so that their quantities are approximately the same when training the unconditional diffusion model. For ablation, we first downsample the non-MLS data so that their quantity is $1\times$, $5\times$, and $10\times$ that of the MLS cases, and then upsample the MLS cases to make them balanced. 
From the results in Fig.~\ref{fig:hist} (b), we find model performance improves with more non-MLS cases incorporated. Increasing non-MLS cases can help train diffusion models and further improve the quality of generated images and extracted feature representations. However, this effect will soon be saturated as the amount of MLS cases is relatively small. This can be a bottleneck for effectively using the non-MLS cases as it is challenging to train unconditional diffusion models with such imbalanced datasets.

\section{Conclusions and Future Work}
In this paper, we propose a novel framework based on deformation field estimation to automatically measure the brain MLS. The labels we are using are sparse which can greatly alleviate the labeling workload. We also propose a semi-supervised learning strategy based on diffusion models which significantly improves the model performance. Experiments on a clinic dataset show our methods can achieve satisfying performance. We also verify that using unlabeled data and non-MLS cases can truly help improve the model's performance. 

Our methods have several limitations. First, the model performance highly relies on pre-trained diffusion models. Training diffusion models with extremely imbalanced data requires great effort. Second, the measurement results exhibit randomness due to noise corruption. Finally, the measurement results are prone to overestimation. Our future work will figure out solutions for these limitations.

%
%
%
%

\end{document}